\documentclass[11pt]{article}

\usepackage[margin=1in]{geometry}
\usepackage{setspace}
\usepackage{amsmath,amssymb}

\usepackage{graphicx}
\graphicspath{{figures/}}

\usepackage{booktabs}
\usepackage{array}

\usepackage{listings}
\usepackage[colorlinks=true,linkcolor=blue,citecolor=blue,urlcolor=blue]{hyperref}

\usepackage[super,comma,sort&compress]{natbib}
\title{TimeTox: An LLM-Based Pipeline for Automated Extraction of Time Toxicity from Clinical Trial Protocols}

\author{
  Saketh Vinjamuri\textsuperscript{1} \and
  Marielle Fis Loperena\textsuperscript{2} \and
  Marie C.\ Spezia\textsuperscript{3} \and
  Ramez Kouzy\textsuperscript{4}
}

\date{March 22, 2026}

\newcommand{\affiliations}{%
\textsuperscript{1}Fairview Hospital, Cleveland Clinic Foundation, Cleveland, OH, USA\\
\textsuperscript{2}The George Washington University School of Medicine and Health Sciences, Washington, DC, USA\\
\textsuperscript{3}University of Missouri-Columbia School of Medicine, Columbia, MO, USA\\
\textsuperscript{4}The University of Texas MD Anderson Cancer Center, Houston, TX, USA
}

\begin{document}

\maketitle

\begin{center}
\small
\affiliations
\end{center}

\vspace{1em}

\begin{abstract}
\textbf{Background:} Time toxicity, the cumulative healthcare contact days from clinical trial participation, is an important but labor-intensive metric to extract from protocol documents. We developed TimeTox, an LLM-based pipeline for automated extraction of time toxicity from Schedule of Assessments tables.

\textbf{Methods:} TimeTox uses Google's Gemini models in three stages: summary extraction from full-length protocol PDFs, time toxicity quantification at six cumulative timepoints for each treatment arm, and multi-run consensus via position-based arm matching. We validated against 20 synthetic schedules (240 comparisons) and assessed reproducibility on 644 real-world oncology protocols. Two architectures were compared: single-pass (vanilla) and two-stage (structure-then-count).

\textbf{Results:} The two-stage pipeline achieved 100\% clinically acceptable accuracy ($\pm$3 days) on synthetic data (MAE 0.81 days) versus 41.5\% for vanilla (MAE 9.0 days). However, on real-world protocols, the vanilla pipeline showed superior reproducibility: 95.3\% clinically acceptable accuracy (IQR $\leq$ 3 days) across 3 runs on 644 protocols, with 82.0\% perfect stability (IQR = 0). The production pipeline extracted time toxicity for 1,288 treatment arms across multiple disease sites.

\textbf{Conclusions:} Extraction stability on real-world data, rather than accuracy on synthetic benchmarks, is the decisive factor for production LLM deployment. Code is available at \url{https://github.com/sakethbuild/TimeTox}.
\end{abstract}


\section{Introduction}

Clinical research generates a wealth of data that remains largely untapped, hidden within complex document structures and dense PDF files. Clinical trial protocols, often exceeding 100 pages, contain detailed specifications of study procedures, visit schedules, and assessment requirements. Within these documents, there exist additional layers of variables that are inherently difficult to identify, subjective in their interpretation, and dispersed across multiple sections without standardized formatting. These characteristics make systematic extraction and analysis at scale exceedingly difficult through manual approaches alone.

One such variable is time toxicity, defined as the cumulative number of healthcare contact days experienced by patients as a result of clinical trial participation~\cite{gupta2022time,shah2024time}. Time toxicity is primarily embodied in the Schedule of Assessments (SoA) table, a dense, multi-page document embedded within clinical trial protocols that specifies all required activities, their timing, and their applicability to each treatment arm. Critically, the SoA table does not follow a unified presentation format across protocols, making it difficult for regulators, trial coordinators, researchers, and patients themselves to understand the per-protocol time commitment of these trials. While we focus here on time, many other clinically relevant variables are similarly embedded within protocol documents and have been the subject of prior extraction efforts~\cite{jin2024matching,agrawal2022llm,tang2023evaluating}.

Manually extracting healthcare contact days from SoA tables requires mapping cycle-based schedules to calendar days, expanding grouped column headers (e.g., ``Cycles 4 to 12'') into individual cycles, distinguishing arm-specific assessments from universal ones, and accumulating unique contact days across cumulative time windows while avoiding double-counting. This process is time-consuming, error-prone, and difficult to standardize across reviewers, as we demonstrate through our own attempts at manual extraction in Section~\ref{sec:groundtruth}.

Recent advances in large language models (LLMs) have demonstrated remarkable capabilities in processing complex structured documents, including medical texts and tabular data~\cite{nori2023gpt4,singhal2023llm}. LLMs can parse visual table layouts, understand domain-specific terminology, and perform multi-step reasoning over structured content, skills directly applicable to SoA table interpretation. However, deploying LLMs for clinical data extraction introduces its own challenges: output variability across runs, sensitivity to prompt design, and the difficulty of validating extractions at scale when ground truth is scarce.

We propose TimeTox, an end-to-end LLM-based pipeline that automates the extraction and calculation of a metric that has traditionally been considered highly subjective. Our contributions include:

\begin{enumerate}
  \item A complete pipeline from raw protocol PDFs to structured time toxicity data, using Gemini models for both document preprocessing and data extraction.
  \item A programmatically generated ground truth dataset of 20 synthetic SoA schedules with clinician-verified time toxicity values, enabling systematic validation.
  \item A comparative evaluation of two extraction architectures, single-pass (vanilla) and two-stage (structure then count), revealing a critical trade-off between synthetic accuracy and real-world stability.
  \item A position-based consensus mechanism that handles LLM-generated arm name instability across runs, enabling robust multi-run aggregation.
  \item Application of the production pipeline to 644 real-world oncology trial protocols, demonstrating feasibility at scale.
\end{enumerate}

\section{Data Collection and Preprocessing}

\subsection{Protocol Acquisition}

We compiled a dataset of 649 clinical trial protocol documents from ClinicalTrials.gov, spanning multiple oncology disease sites including breast, thoracic, gastrointestinal, genitourinary, gynecologic, head and neck, skin, central nervous system, and hematologic malignancies. Protocols were selected based on the availability of full-text documents with Schedule of Assessments (SoA) tables.

\subsection{Summary Extraction}

Full-length clinical trial protocols typically span 100 to 400 pages, of which only 5 to 15 pages contain the SoA table and related scheduling information. To reduce input size and improve extraction accuracy, we developed an automated summary extraction stage using Google's Gemini 2.5 Flash model.

A typical SoA table is presented in a dense grid format, with assessment categories organized as rows and visit time points arranged as columns. Row categories commonly include administrative procedures (informed consent, randomization), clinical assessments (physical examination, vital signs, performance status), laboratory assessments (complete blood count, metabolic panels, liver function tests), cardiac assessments (ECG, echocardiogram), safety monitoring (adverse events, concomitant medications), study treatment administration and accountability, patient reported outcomes (quality of life instruments), and imaging or tumor assessments. Columns represent discrete visit timepoints, including screening, individual cycle days (e.g., C1D1, C1D8, C2D1), grouped cycle ranges (e.g., C4 to C12), end of treatment, and follow-up visits. Cells are marked with indicators such as ``X'' for assessments required of all patients and ``X\textsuperscript{A}'' for assessments specific to a particular treatment arm. A legend section specifies arm-specific visit patterns and cycle structures, while footnotes provide additional details on assessment timing, visit windows, and procedural requirements.

\begin{figure}[htbp]
  \centering
  \includegraphics[width=\textwidth]{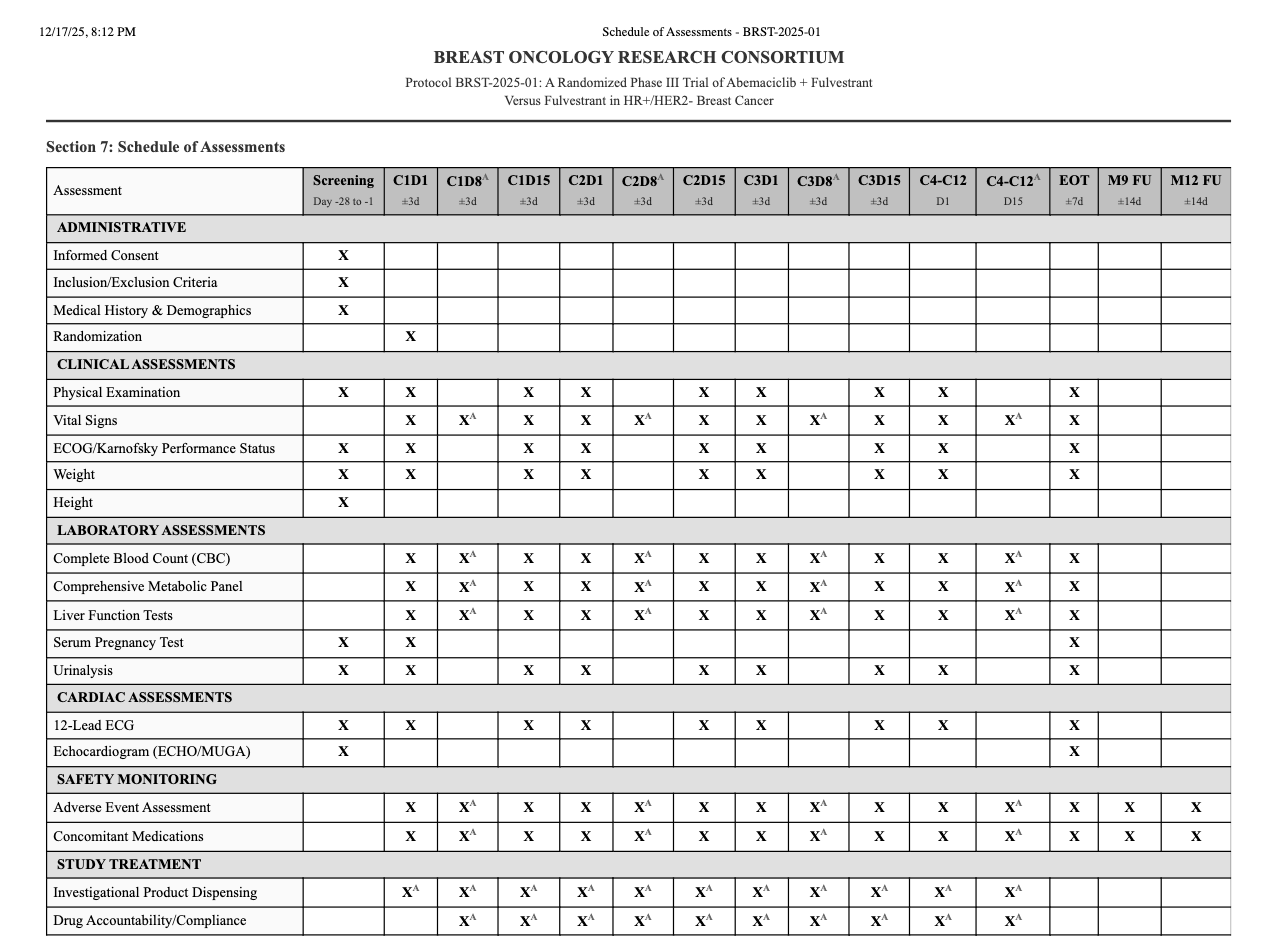}
  \caption{Representative SoA table from a complex synthetic breast oncology protocol (BRST-2025-01) showing two treatment arms with three visit days per cycle.}
  \label{fig:soa_example}
\end{figure}

For each protocol PDF, the summary extraction pipeline:
\begin{enumerate}
  \item Uploads the full PDF to the Gemini API.
  \item Prompts the model to identify pages containing SoA tables, visit schedules, assessment timing, or related scheduling information. The prompt instructs the model to be liberal in page identification to avoid missing relevant content.
  \item Generates a consolidated summary PDF containing: (a) a concise trial summary (study design, treatment arms, duration), and (b) the identified SoA pages plus one surrounding page on each side for context.
\end{enumerate}

\begin{figure}[htbp]
  \centering
  \includegraphics[width=\textwidth]{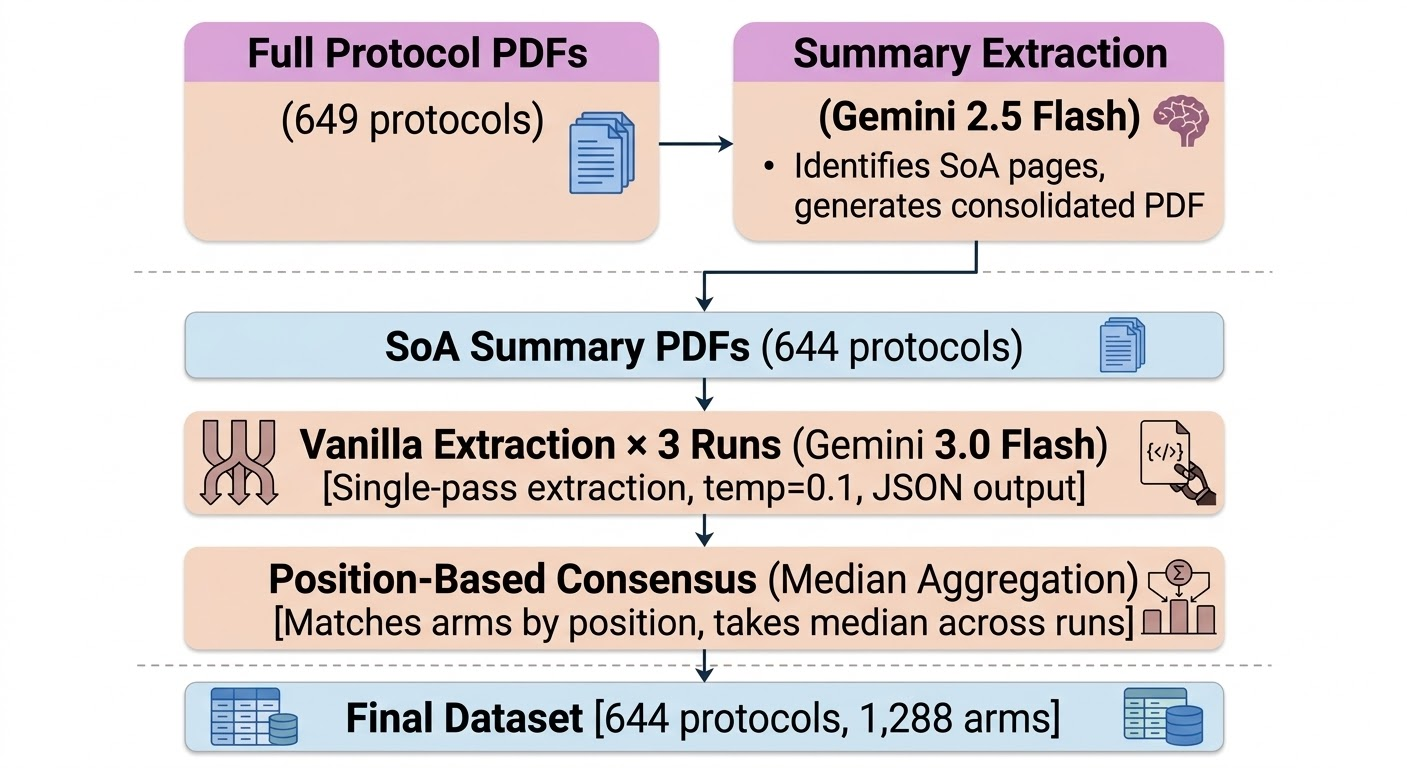}
  \caption{Step-by-step pipeline for processing protocol PDFs via the Gemini API to extract relevant schedules and generate a consolidated summary document.}
  \label{fig:pipeline}
\end{figure}

The summary extraction uses forced JSON output via \texttt{response\_mime\_type="application/json"} with a structured schema, temperature 0.0, and top-p 0.95. Processing time averages 2 to 3 minutes per protocol. Of the 649 protocols processed, summaries were successfully generated for 644, yielding a 99.5\% success rate.

\begin{table}[htbp]
  \centering
  \caption{Dataset Overview}
  \label{tab:dataset}
  \begin{tabular}{ll}
    \toprule
    Characteristic & Value \\
    \midrule
    Total protocols collected & 649 \\
    Protocols with successful summary extraction & 644 \\
    Summary extraction success rate & 99.5\% \\
    Summary extraction model & Gemini 2.5 Flash \\
    Average processing time per protocol & 2 to 3 min \\
    Average SoA pages identified per protocol & 5 to 15 \\
    Disease sites represented & 11 \\
    \bottomrule
  \end{tabular}
\end{table}

\section{Ground Truth Development}
\label{sec:groundtruth}

Before pursuing automated extraction, we first attempted manual extraction of time toxicity from clinical trial protocols. Two independent judges with clinical research experience were tasked with extracting healthcare contact days from a subset of protocols using pre-coded guidelines that specified how to count visits, handle cycle-based schedules, and aggregate contact days across time windows. Despite multiple calibration meetings to align their approaches, the judges were unable to achieve acceptable inter-rater agreement. The high degree of subjectivity inherent in interpreting SoA tables, even with explicit instructions, led to substantial disagreement over visit counting, cycle expansion, and the handling of ambiguous table entries. This difficulty would be further compounded in clinical practice, where time constraints, varying levels of familiarity with protocol conventions, and the sheer volume of protocols under review make consistent manual extraction impractical. The inability to establish reliable human ground truth for real-world protocols motivated our decision to develop programmatically generated synthetic schedules for validation, enabling ground truth to be computed deterministically.

\subsection{Synthetic Schedule Generation}

To enable systematic validation of extraction pipelines, we developed a programmatic generator for synthetic SoA schedules. The generator (\texttt{generate\_schedules.py}, 1,284 lines) creates realistic HTML-rendered schedules with known ground truth values, enabling precise accuracy measurement without the ambiguity inherent in real-world protocol interpretation.

Each synthetic schedule includes a complete SoA table with realistic assessment categories spanning administrative, clinical, laboratory, cardiac, safety, treatment, imaging, and patient-reported outcomes. Every schedule features two treatment arms (intervention and control) with distinct visit schedules, cycle-based timing with configurable cycle lengths of 7, 21, 28, or 35 days, and treatment durations ranging from 2 to 12 months. Schedules also incorporate a legend section specifying arm-specific visit patterns, footnotes, headers, and formatting consistent with real clinical trial protocols, as well as disease-specific assessments appropriate to each tumor type.

We generated 20 synthetic schedules spanning 8 oncology disease categories (breast, thoracic, gastrointestinal, genitourinary, head \& neck, melanoma, gynecologic, and sarcoma) with three complexity levels: simple ($n$=5, single visit day per cycle), moderate ($n$=10, two visit days per cycle), and complex ($n$=5, three visit days per cycle). Five distinct visual styles were used to test robustness to formatting variation. Schedules incorporated diverse treatment modalities, including systemic therapy, radiation, and surgery.

\subsection{Ground Truth Calculation}

Ground truth healthcare contact days were computed deterministically from each schedule's configuration parameters. Treatment visit days were first mapped from cycle-based schedules to absolute calendar days. Imaging days were then added at protocol-specified intervals, and end-of-treatment (EOT) and follow-up visits at 9 and 12 months were included. Finally, unique calendar days were counted at each cumulative time point (screening, 1, 3, 6, 9, and 12 months). This yielded 40 arms across 6 time points, yielding 240 ground-truth comparisons.

\begin{table}[htbp]
  \centering
  \caption{Synthetic Schedule Characteristics}
  \label{tab:synthetic}
  \begin{tabular}{ll}
    \toprule
    Characteristic & Value \\
    \midrule
    Total schedules & 20 \\
    Total arms & 40 (20 intervention, 20 control) \\
    Total ground truth comparisons & 240 \\
    Disease categories & 8 \\
    Complexity: Simple / Moderate / Complex & 5 / 10 / 5 \\
    Cycle lengths & 7, 21, 28, 35 days \\
    Treatment durations & 2 to 12 months \\
    Visual style variants & 5 \\
    Treatment modalities & Systemic, Radiation, Surgery \\
    \bottomrule
  \end{tabular}
\end{table}

\subsection{Ground Truth Verification}

The programmatically generated ground-truth values were independently verified by multiple reviewers, including a clinician expert who assessed the clinical plausibility of the synthetic schedules and confirmed the accuracy of the calculated contact days. This verification ensured that both the schedule designs and the ground-truth calculations accurately reflected real-world clinical trial visitation patterns.

\section{LLM Pipeline Design}

We evaluated two extraction architectures, both using Google's Gemini 3.0 Flash (model ID: gemini-3-flash-preview, accessed January to February 2026) as the base model with temperature 0.1 to balance accuracy and determinism. The summary extraction stage (Section~2.2) used Gemini 2.5 Flash (model ID: gemini-2.5-Flash-preview-04-17, accessed December 2025 to January 2026). All model interactions used the Google Generative AI API.

\subsection{Vanilla Pipeline (Primary)}

The vanilla pipeline performs single-pass extraction, providing the model with a summary SoA PDF and a structured prompt that requests direct quantification of healthcare contact days. The prompt (76 lines of instruction, excluding the JSON output template) specifies:

\begin{itemize}
  \item \textbf{Definitions:} Healthcare contact day definition, six cumulative time windows (screening through 12 months), and the instruction to count each unique calendar day once, regardless of the number of assessments.
  \item \textbf{Extraction rules:} Five key rules covering arm identification, cycle-to-calendar mapping, grouped column expansion, unique day counting, and arm-specific visit handling (e.g., superscript markers like X\textsuperscript{A}).
  \item \textbf{Category definitions:} Four categories of healthcare contact (core treatment, imaging/diagnostics, labs, and clinic visits), with the note that a single day may span multiple categories.
  \item \textbf{Output format:} Strict JSON schema specifying arm name, intervention type, healthcare contact days at each timepoint, category breakdown, and extraction notes (cycle length, treatment duration, visit pattern, disease).
\end{itemize}

The model receives the PDF as a visual input and produces a JSON response. Three retry attempts with exponential backoff handle transient API failures.

\begin{lstlisting}[caption={Condensed excerpt of the vanilla extraction prompt.},label={lst:prompt}]
You are an expert clinical trial protocol analyst.
Your task is to extract healthcare contact days
from the Schedule of Assessments (SoA) table.

## DEFINITIONS
**Healthcare Contact Day**: Any calendar day requiring
in-person contact with the healthcare system.
Count each unique calendar day ONCE.

**Time Windows** (cumulative):
- screening: All pre-treatment visits
- 1_month: Day 1 through Day 30
- 3_months: Day 1 through Day 90
...
- 12_months: Day 1 through Day 365

## EXTRACTION RULES
1. Identify all treatment arms
2. Map cycle structure to calendar days
3. CRITICAL: Expand grouped columns (e.g., "C4-C12")
4. Count unique visit days
5. Handle arm-specific visits (superscripts)

## OUTPUT FORMAT
Return ONLY valid JSON: [{"arm_name": "...",
  "healthcare_contact_days": {...}, ...}]
\end{lstlisting}

\subsection{Two-Stage Pipeline (Alternative)}

The two-stage pipeline decomposes extraction into two sequential LLM calls:

\textbf{Stage 1, Structure Extraction:} The model extracts the structural parameters of the SoA table, including: cycle length in days, visit days per cycle for each arm, treatment duration, special visit information (screening days, EOT visit, follow-up visits), and disease type. The output is a structured JSON object capturing the schedule's ``blueprint'' without performing any counting. This stage uses forced JSON output (\texttt{response\_mime\_type="application/json"}).

\textbf{Stage 2, Count Calculation:} The extracted structure from Stage 1, formatted as JSON, is provided alongside the original PDF as context for a second LLM call. This call performs the arithmetic: mapping cycles to calendar days, applying visit patterns, and accumulating unique contact days at each timepoint. The prompt includes explicit calculation rules (e.g., ``Total cycles = floor(treatment\_duration\_months $\times$ 30 / cycle\_length\_days)'').

The rationale for decomposition was to separate the perceptual task (reading the table) from the computational task (counting days), hypothesizing that this division would reduce errors. As described in Section~\ref{sec:stability}, while this approach achieved superior accuracy on synthetic data, it exhibited unacceptable variability on real-world protocols.

\section{Validation on Synthetic Data}

\subsection{Evaluation Metrics}

We evaluated both pipelines using three metrics computed across all 240 ground-truth comparisons (20 schedules, 2 arms, 6 time points). Exact Match Accuracy measures the percentage of extractions that match the ground truth exactly, providing the strictest assessment of pipeline precision. Clinically Acceptable Accuracy (within 3 days) captures the percentage of extractions falling within 3 days of the ground truth, a threshold chosen because differences of fewer than 3 healthcare contact days over a 12-month period are unlikely to be clinically meaningful or to alter comparative conclusions between treatment arms. Mean Absolute Error (MAE) quantifies the average magnitude of extraction errors in days, providing a continuous measure of pipeline performance.

It is important to note that the variables extracted by our pipeline represent protocol-mandated time toxicity, which constitutes a best-case scenario for the actual time burden experienced by patients. Real-world healthcare contact may exceed protocol requirements due to unscheduled visits, adverse event management, and logistical factors. Our extraction, therefore, provides a lower-bound estimate that is nonetheless valuable for systematic cross-protocol comparison.

In designing our evaluation framework, we deliberately prioritized low variability across runs over the possibility of systematic shifts in absolute values. We accepted that the pipeline might consistently overestimate or underestimate healthcare contact days by a fixed margin. However, if such a systematic shift exists, consistency across runs is critical: a pipeline that consistently overestimates by a small amount still produces values that can be meaningfully compared across protocols and treatment arms in downstream analysis. Conversely, a pipeline with high inter-run variability, even if occasionally exact, cannot support reliable comparative conclusions. This reasoning guided our emphasis on stability testing (Section~\ref{sec:stability}) as the primary criterion for production deployment.

\subsection{Results}

The two-stage pipeline substantially outperformed the vanilla pipeline on all synthetic accuracy metrics, achieving perfect clinically acceptable accuracy. However, as Section~\ref{sec:stability} demonstrates, these synthetic metrics did not predict real-world performance.

\begin{table}[htbp]
  \centering
  \caption{Validation Results on 20 Synthetic Schedules (240 Comparisons)}
  \label{tab:validation}
  \begin{tabular}{lccc}
    \toprule
    Pipeline & Exact Match & Clinically Acceptable Accuracy ($\pm$3d) & MAE (days) \\
    \midrule
    Vanilla (single-pass) & 0.3\% & 41.5\% & 9.0 \\
    Two-Stage (structure $\rightarrow$ count) & 29.2\% & 100.0\% & 0.81 \\
    \bottomrule
  \end{tabular}
\end{table}

\subsubsection{Error Distribution Analysis of the Vanilla Pipeline}
\label{sec:error_distribution}

To characterize the nature of the vanilla pipeline's errors on synthetic data, we analyzed the signed error (extracted $-$ ground truth) across all 240 comparisons. Overall, the median signed error was +4.0 days (mean +6.9 days), indicating a systematic tendency toward overcounting. The pipeline overcounted in 65.4\% of comparisons, undercounted in 34.2\%, and achieved an exact match in only 0.4\%.

Table~\ref{tab:error_dist} presents the error distribution by time window. At screening, the pipeline consistently undercounted by 1 day (90\% undercount rate), likely due to misidentifying the number of pre-treatment visits. For longer time windows, overcounting dominated: 75 to 80\% of comparisons showed positive errors at 3 to 12 months, with the median signed error increasing from +2.0 days at 1 month to +11.5 days at 12 months. This pattern reflects cumulative error propagation as cycle-to-day mapping errors compound over longer periods.

\begin{table}[htbp]
  \centering
  \caption{Vanilla Pipeline Error Distribution on Synthetic Data by Time Window (240 Comparisons)}
  \label{tab:error_dist}
  \begin{tabular}{lcccccc}
    \toprule
    Time Window & Median Error & Mean Error & MAE & \% Overcount & \% Undercount & \% Exact \\
    \midrule
    Screening & $-$1.0 & $-$0.2 & 1.6 & 10.0\% & 90.0\% & 0.0\% \\
    1 Month & +2.0 & +1.9 & 2.2 & 77.5\% & 20.0\% & 2.5\% \\
    3 Months & +5.5 & +4.9 & 5.8 & 80.0\% & 20.0\% & 0.0\% \\
    6 Months & +10.0 & +9.1 & 11.3 & 75.0\% & 25.0\% & 0.0\% \\
    9 Months & +11.5 & +11.9 & 14.1 & 75.0\% & 25.0\% & 0.0\% \\
    12 Months & +11.5 & +14.0 & 16.3 & 75.0\% & 25.0\% & 0.0\% \\
    \bottomrule
  \end{tabular}
\end{table}

Error magnitude also varied by schedule complexity. Complex schedules (3 visit days per cycle) had the highest mean absolute error (12.8 days) and the strongest overcounting tendency (85\% overcount rate), compared with 7.1 days MAE for moderate and 7.3 days for simple schedules. This suggests that the vanilla pipeline's single-pass approach is particularly challenged by multi-visit-day cycles, where grouped column expansion requires additional arithmetic steps.

\begin{figure}[htbp]
  \centering
  \includegraphics[width=\textwidth]{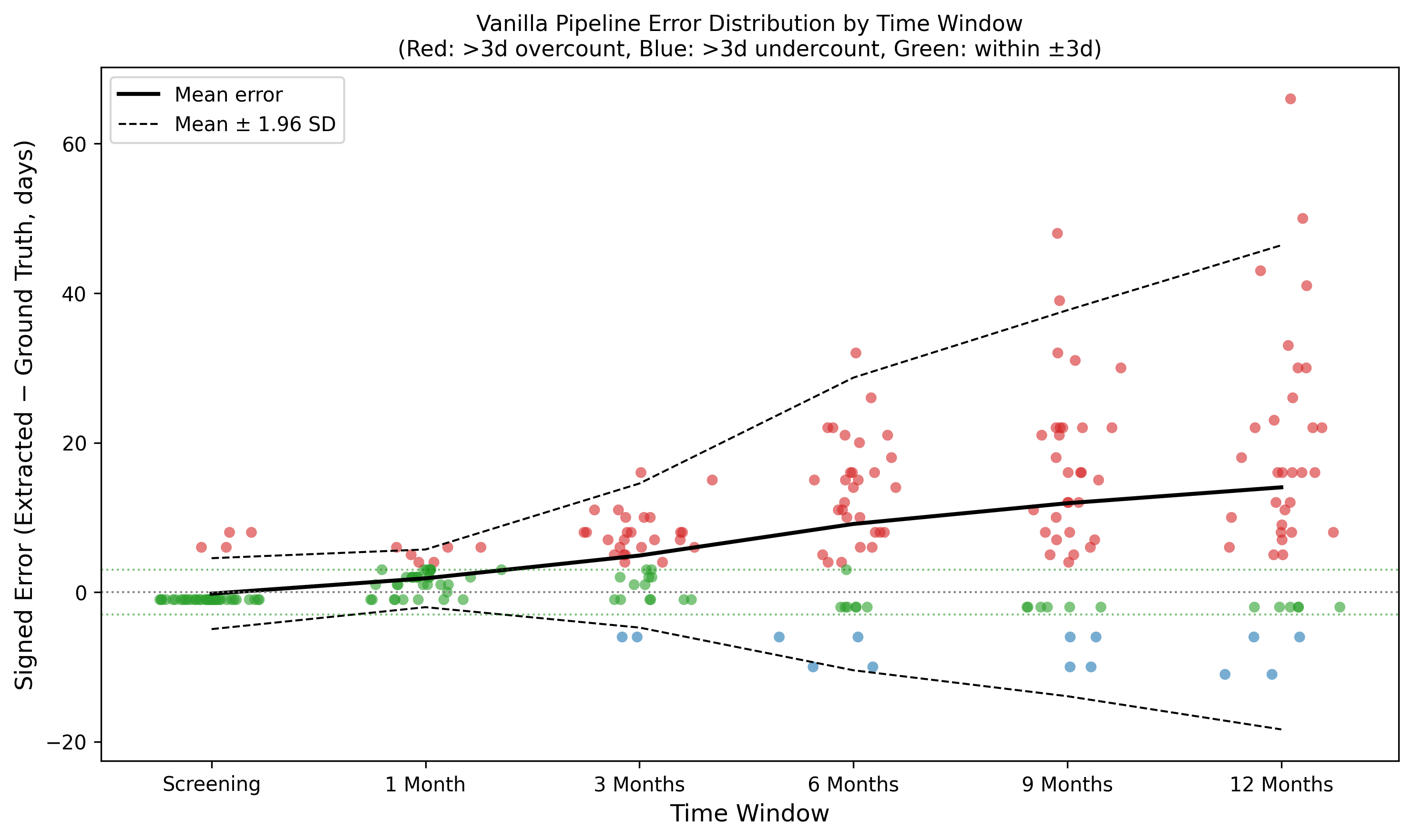}
  \caption{Vanilla pipeline signed error distribution by time window on 20 synthetic schedules (240 comparisons). Each dot represents one arm-timepoint comparison. Red: overcount $>$3 days; blue: undercount $>$3 days; green: within $\pm$3 days. Solid line: mean error; dashed lines: mean $\pm$ 1.96 SD (limits of agreement).}
  \label{fig:bland_altman}
\end{figure}

\begin{figure}[htbp]
  \centering
  \includegraphics[width=0.85\textwidth]{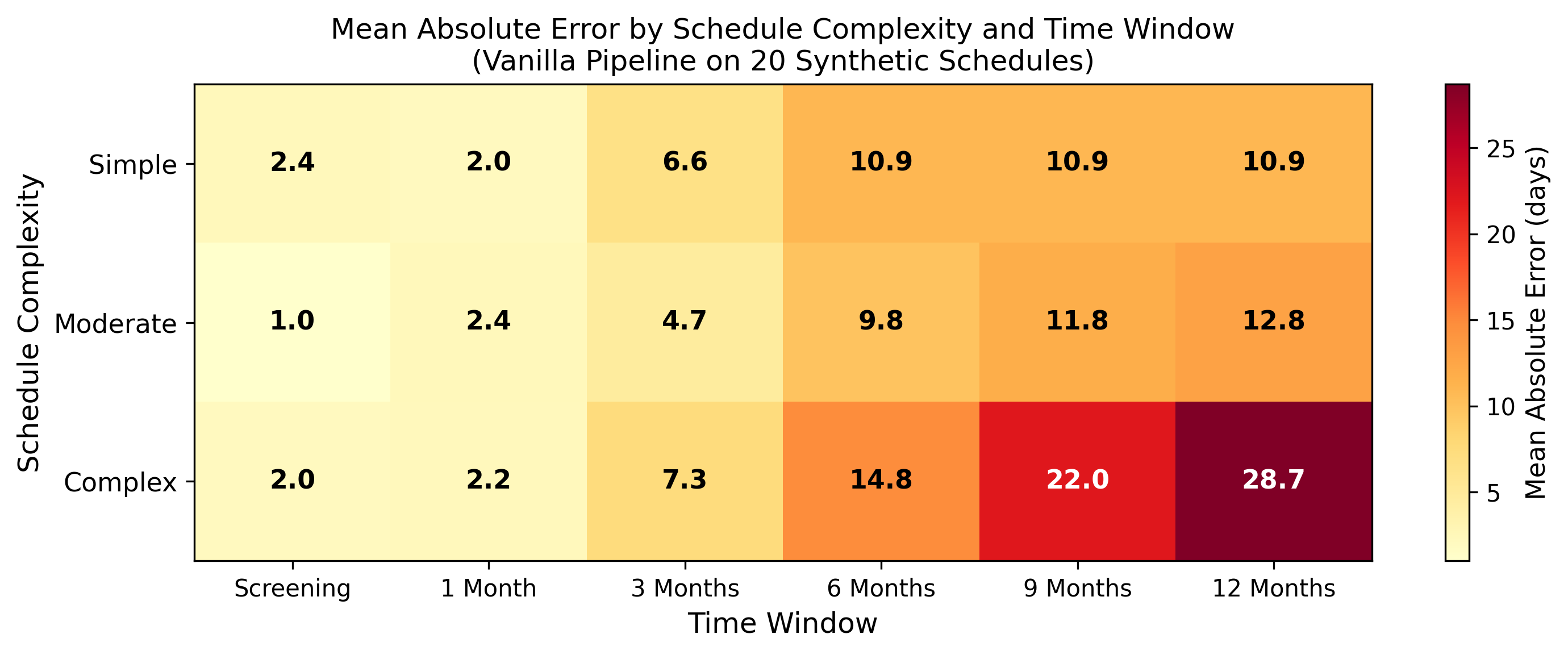}
  \caption{Mean absolute error (days) by schedule complexity and time window for the vanilla pipeline on synthetic data. Errors are lowest at screening and increase with both time window duration and schedule complexity.}
  \label{fig:error_heatmap}
\end{figure}

The vanilla pipeline's low synthetic accuracy is attributable to the inherent difficulty of performing multi-step arithmetic (cycle mapping, grouped column expansion, cumulative counting) in a single LLM call. The model frequently miscounts visits in grouped columns or makes errors in cycle-to-day conversion. In contrast, the two-stage pipeline benefits from explicit structural extraction followed by guided calculation, reducing the cognitive load on each individual call.

\section{Stability and Reproducibility Testing}
\label{sec:stability}

\subsection{Rationale for Stability Testing}

While synthetic accuracy measures correctness against known ground truth, production deployment requires a fundamentally different property: reproducibility. If the same protocol yields substantially different time toxicity values across independent extraction runs, the resulting dataset cannot support reliable downstream analysis, regardless of whether any individual run happens to be accurate.

As discussed in Section~5.1, our evaluation framework accepts the possibility of systematic shifts in extracted values but demands consistency. A pipeline that reliably produces values within a stable range, even if that range is offset from the true value, enables valid comparative analysis across protocols, treatment arms, and disease sites. The key insight is that for most downstream research questions (e.g., does immunotherapy impose more time toxicity than chemotherapy?), relative comparisons matter more than absolute accuracy. What undermines such analysis is not systematic bias but rather high variance, where the same protocol might yield 30 contact days on one run and 90 on the next. We therefore conducted systematic stability testing on real-world protocols to determine which pipeline architecture could reliably support large-scale comparative analysis.

\subsection{Phase 1: Pilot Stability Study (5 Runs, 100 Protocols)}

We first compared both pipelines on a pilot set of real-world protocols. Using 5 real-world summary PDFs, we ran each pipeline 5 times and measured the inter-quartile range (IQR) of the 12-month healthcare contact day count across runs for each arm.

The two-stage pipeline exhibited substantial inter-run variability on real protocols. For example, one protocol arm produced values of [94, 473, 473, 540, 543] across 5 two-stage runs (IQR = 258.0), while the same arm under the vanilla pipeline produced [99, 105, 106, 106, 106] (IQR = 4.0). Another protocol yielded two-stage values of [27, 53, 53, 53, 53] (IQR = 13.0) versus vanilla values of [8, 8, 8, 8, 8] (IQR = 0.0).

This discrepancy, namely high synthetic accuracy but high real-world variability for the two-stage pipeline, reflects a fundamental challenge: real-world SoA tables are far more heterogeneous than synthetic schedules. Ambiguous column headers, inconsistent formatting, multi-page tables, and complex footnotes create variability in interpretation, which is amplified by the two-stage pipeline's reliance on accurate structural extraction in Stage~1.

We then extended stability testing to 100 real-world protocols with 5 independent vanilla runs.

\begin{table}[htbp]
  \centering
  \caption{Vanilla Pipeline Stability: 5 Runs on 100 Real-World Protocols}
  \label{tab:stability_pilot}
  \begin{tabular}{lcc}
    \toprule
    Stability Category & Arms & Percentage \\
    \midrule
    Perfect Stability (IQR = 0) & 417 & 77.2\% \\
    Clinically Acceptable Accuracy (IQR $\leq$ 3) & 506 & 93.7\% \\
    High Variance (IQR $>$ 3) & 34 & 6.3\% \\
    \midrule
    Total arms analyzed & 540 & \\
    \bottomrule
  \end{tabular}
\end{table}

\subsection{Phase 2: Full-Scale Stability (3 Runs, 644 Protocols)}

Based on the pilot results, we selected the vanilla pipeline for production deployment and ran 3 independent extractions on all 644 successfully preprocessed protocols.

\begin{table}[htbp]
  \centering
  \caption{Vanilla Pipeline Stability: 3 Runs on 644 Real-World Protocols}
  \label{tab:stability_full}
  \begin{tabular}{lcc}
    \toprule
    Stability Category & Arms & Percentage \\
    \midrule
    Perfect Stability (IQR = 0) & 2,172 & 82.0\% \\
    Clinically Acceptable Accuracy (IQR $\leq$ 3) & 2,524 & 95.3\% \\
    High Variance (IQR $>$ 3) & 125 & 4.7\% \\
    \midrule
    Total arms analyzed & 2,649 & \\
    \bottomrule
  \end{tabular}
\end{table}

The improvement from 93.7\% to 95.3\% clinically acceptable accuracy when moving from 5 runs on 100 protocols to 3 runs on 644 protocols likely reflects both the natural variance reduction with fewer runs and the broader protocol sample smoothing out edge cases.

\subsection{Stability Metrics}

We define clinically acceptable accuracy as follows:

\begin{itemize}
  \item \textbf{Perfect Stability (IQR = 0):} All runs produced identical 12-month contact day counts for this arm.
  \item \textbf{Clinically Acceptable Accuracy (IQR $\leq$ 3):} The interquartile range of 12-month counts is at most 3 days, a threshold below which differences are not clinically meaningful.
  \item \textbf{High Variance (IQR $>$ 3):} Substantial disagreement across runs, flagging arms for potential review.
\end{itemize}

\section{Production Pipeline and Consensus}

\subsection{Pipeline Architecture}

Based on the validation and stability results, we adopted the vanilla pipeline with 3-run consensus as the production configuration.

\begin{figure}[htbp]
  \centering
  \includegraphics[width=\textwidth]{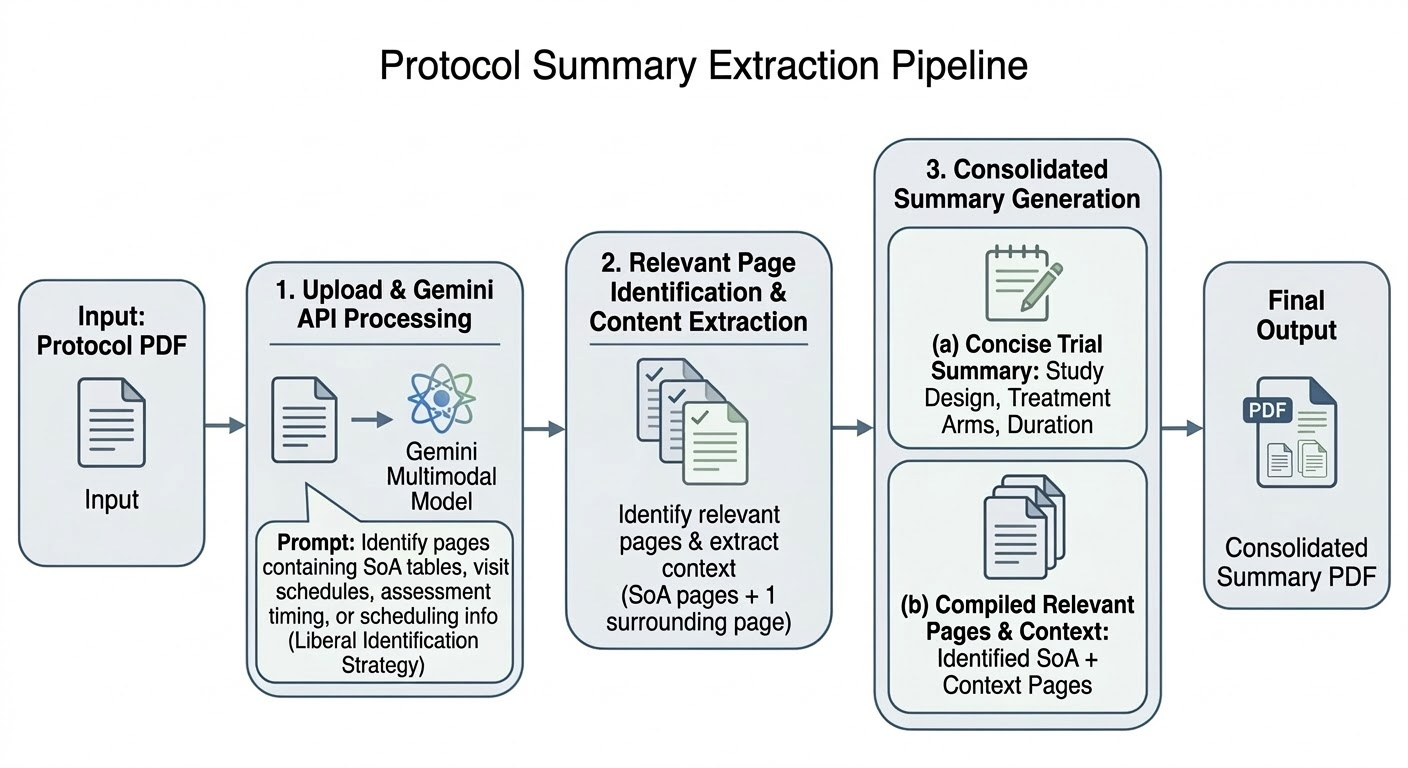}
  \caption{TimeTox Production Pipeline Architecture.}
  \label{fig:production}
\end{figure}

\subsection{Position-Based Consensus Mechanism}

A key challenge in multi-run LLM extraction is arm name instability: the same treatment arm may be labeled differently across runs (e.g., ``Arm A (Pembrolizumab + Chemotherapy)'' vs.\ ``Pembrolizumab + Chemotherapy Arm'' vs.\ ``Treatment Arm A: Pembro/Chemo''). Standard name-based matching fails in this scenario.

We address this through position-based matching:
\begin{enumerate}
  \item Within each run, arms are grouped by (filename, intervention\_type).
  \item Within each group, arms are sorted by their 12-month healthcare contact day count.
  \item A positional index (pos\_idx) is assigned based on this sorted order.
  \item Consensus is computed by grouping across runs by (filename, intervention\_type, pos\_idx) and taking the median of each timepoint.
\end{enumerate}

This approach leverages the insight that, even when arm names vary, the relative ordering of arms by time-toxicity burden is stable. The intervention type (intervention vs.\ control) is normalized during preprocessing (e.g., ``placebo'' $\rightarrow$ ``control'', ``active\_comparator'' $\rightarrow$ ``intervention'').

\begin{table}[htbp]
  \centering
  \caption{Production Consensus Dataset Statistics}
  \label{tab:production}
  \begin{tabular}{ll}
    \toprule
    Metric & Value \\
    \midrule
    Protocols processed & 644 \\
    Total consensus arms & 1,288 \\
    Arms with 3 concordant runs & majority \\
    Extraction model & Gemini 3.0 Flash \\
    Extraction temperature & 0.1 \\
    Extraction runs per protocol & 3 \\
    Consensus method & Median aggregation \\
    Average processing time per extraction & $\sim$4 min \\
    \bottomrule
  \end{tabular}
\end{table}

\subsubsection{Cross-Arm Position Stability}
\label{sec:crossarm}

The position-based consensus mechanism assumes that the relative ordering of arms by 12-month contact day count is stable across runs. If two arms within the same intervention type have similar contact days, their ordering could swap across runs, leading to cross-arm contamination in the consensus. We analyzed the frequency of such position swaps across the 5 stability runs on 100 real-world protocols.

Of 202 unique (filename, intervention\_type) groups across all runs, 24 (11.9\%) contained more than one arm of the same intervention type, making them susceptible to position swaps. Among these 24 groups, 3 (12.5\%) exhibited potential position swaps where the 12-month contact day values at adjacent positions overlapped across runs. At the arm-pair level, 7 of 16 adjacent same-type arm pairs (43.8\%) had 12-month contact day values within 3 days of each other, indicating that close-value pairs are common when multiple arms of the same type exist.

The low overall swap rate (3/24 groups affected) suggests that position-based consensus is robust for the majority of protocols. However, the high close-pair rate among multi-arm groups indicates that for the subset of protocols with multiple arms of similar time toxicity burden, position-based matching may introduce some cross-arm mixing. The impact is mitigated by the median aggregation step, which dampens the effect of any single swapped run. Nonetheless, this limitation should be considered when interpreting results for protocols with multiple arms of the same intervention type and similar contact day counts.

\subsection{Cost and Runtime}

The full production pipeline cost is dominated by API calls to the Gemini models. Summary extraction (Gemini 2.5 Flash) processes each protocol in approximately 2 to 3 minutes. Each vanilla extraction run (Gemini 3.0 Flash) takes approximately 4 minutes per protocol. With 3 runs across 644 protocols, the total extraction time is approximately 128 hours of sequential processing (which can be reduced through parallelization). The Gemini API pricing makes the total cost well under \$100 for the complete dataset, making the approach economically viable for large-scale studies.

\section{Application}

To demonstrate the pipeline's utility, we applied the production consensus dataset to a landscape analysis of time toxicity across oncology clinical trials. The full clinical analysis, including descriptive statistics, comparative analyses across disease sites and treatment modalities, temporal trends, multivariable regression, and component analysis, is presented in a companion paper.

Briefly, the pipeline enabled analysis of time toxicity patterns across 644 protocols spanning 11 disease sites, multiple treatment modalities (chemotherapy, immunotherapy, targeted therapy, endocrine therapy, radiation, and procedures), and trial start years from 1994 to 2021. The median 12-month time toxicity across all intervention arms was quantified, and systematic differences by disease site, treatment type, and funding source were characterized. These analyses would be infeasible to perform manually at this scale.

\section{Discussion}

\subsection{Accuracy vs.\ Stability Trade-off}

The central finding of this work is the critical distinction between synthetic accuracy and real-world stability. The two-stage pipeline achieved perfect clinically acceptable accuracy on synthetic schedules but exhibited unacceptable variability on real-world protocols. Conversely, the vanilla pipeline showed modest synthetic accuracy but remarkable stability, with 95.3\% of arms having an IQR $\leq$ 3 days across independent runs.

This divergence has a straightforward explanation. Synthetic schedules have an unambiguous structure: clear column headers, explicit legends, and well-defined visit patterns. The two-stage pipeline excels when the structural extraction step can parse this structure cleanly. Real-world protocols, however, contain multi-page tables with complex footnotes, ambiguous column groupings, inconsistent notation, and formatting artifacts. In this environment, small variations in structural extraction at Stage~1 cascade into large differences in counts at Stage~2. The vanilla pipeline, by avoiding explicit structural decomposition, is more robust to these sources of ambiguity because it treats extraction as a holistic task, reducing the number of failure points.

\subsection{Design Decisions}

\textbf{Temperature selection:} We used temperature of 0.1 throughout the extraction. Lower temperatures (0.0) with forced JSON output were tested during development, but did not consistently improve stability, while they occasionally caused the model to produce malformed outputs. Temperature 0.1 provided a balance between determinism and output quality.

\textbf{Prompt design:} The vanilla prompt was iteratively refined. Key additions included an explicit instruction to expand grouped columns (``CRITICAL: Expand grouped columns''), a specification that time windows are cumulative, and a category breakdown requirement. Each addition addressed a systematic error pattern observed during development.

\textbf{Three-run consensus:} Three runs were chosen as a pragmatic balance between stability and cost. Increasing from 3 to 5 runs would provide a marginal improvement in stability (as evidenced by the 93.7\% vs.\ 95.3\% comparison) at a 67\% increase in cost.

\subsection{Limitations}

Several limitations should be acknowledged. First, all extraction relies on a single LLM family, Google's Gemini models. While Gemini 3.0 Flash demonstrated good performance, a comparative evaluation against other LLM families, such as GPT-4 and Claude, was not conducted using formal benchmarking. Second, clinically acceptable accuracy was formally validated only against synthetic schedules; real-world validation relied on stability testing and informal spot checking against manually reviewed protocols rather than a comprehensive gold standard. Third, the vanilla pipeline may systematically under- or overcount healthcare contact days, and while the consensus mechanism reduces random error, it cannot correct consistent directional bias across runs. Fourth, extremely complex protocols, such as multi-phase adaptive designs with dose-escalation cohorts, may exceed the pipeline's extraction capability, contributing to the 4.7\% of arms with high variance. Finally, the pipeline captures contact days up to 12 months; longer-term follow-up visits beyond this window are not captured.

\subsection{Comparison to Related Work}

LLM-based information extraction from medical documents is an active area of research. Agrawal et al.\cite{agrawal2022llm} demonstrated that large language models can serve as effective few-shot extractors of clinical information, achieving competitive performance on tasks such as extracting medication attributes and clinical trial criteria from unstructured text. Tang et al.\cite{tang2023evaluating} evaluated LLMs on medical evidence summarization, showing both the promise and limitations of current models in distilling complex medical information. In the clinical trial domain specifically, Jin et al.\cite{jin2024matching} explored matching patients to clinical trials using LLMs, demonstrating the feasibility of automated reasoning over trial eligibility criteria. More broadly, Singhal et al.\cite{singhal2023llm} showed that LLMs encode substantial clinical knowledge, while Nori et al.\cite{nori2023gpt4} documented the capabilities of GPT-4 on medical challenge problems, establishing the foundation for clinical document understanding tasks.

Our work extends this line of research to a specific, previously unautomated task, time toxicity quantification from SoA tables, that requires both visual table parsing and multi-step arithmetic reasoning. Unlike prior extraction tasks that operate primarily on unstructured narrative text, our pipeline must interpret complex tabular structures with visual formatting cues, hierarchical column headers, and cross-referenced footnotes. This introduces challenges distinct from those addressed in prior work, particularly the need for spatial reasoning over table layouts and deterministic multi-step calculation from extracted structural parameters.

The stability-focused evaluation framework we employ is less common in the LLM extraction literature, which typically emphasizes accuracy against gold standards. Recent work on LLM reliability in clinical settings~\cite{clusmann2023future} has highlighted the importance of output consistency, particularly when extracted data feeds into downstream statistical analyses. Our finding that stability on heterogeneous real-world data is a better predictor of production utility than accuracy on controlled benchmarks may generalize to other LLM-based extraction tasks, especially those involving complex structured documents where interpretation ambiguity is inherent.

\section{Conclusion}

We presented TimeTox, an LLM-based pipeline for automated extraction of time toxicity from clinical trial protocol SoA tables. The pipeline combines Gemini-based summary extraction, vanilla single-pass time toxicity extraction, and position-based multi-run consensus to produce stable, scalable quantification of healthcare contact days.

Our evaluation revealed a critical insight for LLM deployment in clinical data extraction: performance on controlled synthetic benchmarks does not predict real-world reliability. The two-stage pipeline's 100\% synthetic accuracy was undermined by high variability on real protocols, while the vanilla pipeline's modest 41.5\% synthetic accuracy translated to 95.3\% clinically acceptable accuracy in production, the metric that ultimately matters for downstream analysis.

The production pipeline was successfully applied to 644 real-world oncology trial protocols, enabling large-scale quantification of time toxicity that would be infeasible through manual extraction. Future work includes expanding ground-truth validation using manually annotated real-world protocols, evaluating multi-model ensemble approaches, and extending the framework to other structured clinical trial data extraction tasks.

\subsection*{Code Availability}

The complete pipeline code, synthetic schedule generator, and analysis scripts are available at \url{https://github.com/sakethbuild/TimeTox}.

\bibliography{references}

\end{document}